\documentclass{WileyMSP-template}
\usepackage{comment}
\usepackage{hyperref}
\usepackage{threeparttable}
\usepackage{xcolor}
\usepackage{booktabs}
\usepackage{lineno}
\usepackage{amsmath}

\begin{document}

\pagestyle{fancy}
\rhead{\includegraphics[width=2.5cm]{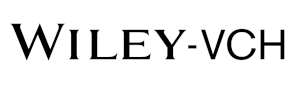}}

\title{
Auditory–Tactile Congruence for Synthesis of Adaptive \\ Pain Expressions  in RoboPatients
}
\maketitle

\author{Saitarun Nadipineni $^1$*}
\author{Chapa Sirithunge $^2$*}
\author{Yue Xie $^2$}
\author{Fumiya Iida $^2$}
\author{Thilina Dulantha Lalitharatne $^1$}

\dedication{}


\begin{affiliations}
$^1$School of Engineering and Materials Science, Queen Mary University of London, London E1 4NS, UK\\ 

$^2$Department of Engineering, University of Cambridge, Cambridge CB2 1PZ, UK\\
Correspondence: s.nadipineni@qmul.ac.uk

$^*$Authors contributed equally to this work.

\end{affiliations}

\keywords{Robot-assisted Training, Synthetic Pain, Palpation, Medical Training Simulators, Human-Robot Interaction}

\begin{abstract}

Misdiagnosis can lead to delayed treatments and harm. Robopatients provide a controlled method for training and evaluating clinicians in rare, subtle, and complex cases, thereby reducing diagnostic errors. We present a study exploring auditory-tactile congruence for the synthesis of adaptive pain expressions in robopatients. In this work, the robopatient functions as an adaptive intermediary, capable of synthesizing vocal pain expressions in response to tactile stimuli generated during palpation. Using an abdominal phantom, robopatient captures and processes haptic input via an internal palpation-to-pain mapping model. To evaluate perceptual congruence between palpation and the corresponding auditory output, we conducted a study involving 7680 trials across 20 participants, who evaluated pain intensity through sound. Results show amplitude and pitch significantly influence agreement with the robot’s pain expressions, irrespective of pain sounds. Stronger palpation forces elicited stronger agreement, aligning with psychophysical patterns. The study revealed that pitch and amplitude are key in pain sound perception, with pitch being more influential. These acoustic features shape how well the sound matches the applied force during palpation, impacting perceived realism. This approach lays the groundwork for high-fidelity robopatients in clinical education and diagnostic simulation, serving as a platform to explore high-dimensional concepts such as pain through robots.

\end{abstract}

\section{Introduction}

Misdiagnosis is a critical global health issue \cite{singh2017global}, resulting in an estimated 795,000 cases of permanent harm or death annually in the United States alone \cite{NewmanToker2024}. The economic burden of diagnostic errors in the U.S. healthcare system is estimated to exceed \$100 billion per year \cite{ball2015improving}. These figures highlight the urgent need for effective training tools that improve diagnostic accuracy and reduce preventable harm. Furthermore, the UK government’s 10-year plan published in 2025, highlights the use of assistive robotic technologies and aims to reduce the need for hospital-based diagnostics \cite{DHSC_2025_FitForTheFuture}. In this context, robotic patients or robopatients provide a high-fidelity, standardized medical training simulators for healthcare professionals. Unlike human actors, robopatients can consistently simulate rare or subtle conditions, allowing for repeated exposure and skill refinement. 

\vspace{3mm}

Palpation is a critical diagnostic skill where clinicians rely heavily on interpreting subtle pain responses. Synthetic pain generation, combining facial expressions, vocalizations, and haptic responses, is especially crucial in training scenarios such as palpation, where visual or auditory pain cues guide clinical judgment \cite{lalitharatne2022face, protpagorn2023vocal}. By enabling controlled, vocal pain feedback in response to tactile input, robopatients help practitioners develop better sensitivity to patients' pain responses. This approach enhances diagnostic precision and supports the development of empathy-driven care. However, accurately recognizing and interpreting another person's pain based on vocal pain information remains underexplored. This gap arises from challenges in defining pain within diverse ontological frameworks \cite{raja2020revised, m2020deep}, as well as the limited understanding of pain's underlying physiology \cite{ellison2017physiology}. 

\vspace{3mm}

Pain is often communicated through facial expressions \cite{prkachin2009assessing}, which are influenced not only by physiological states but also by social and contextual factors \cite{vlaeyen2009threat, craig2015social}. Despite its inherently interdisciplinary nature, spanning neuroscience, psychology, social contexts, and computer science, the study of pain has largely lacked comprehensive cross-disciplinary approaches. This limitation highlights the need for novel exploratory paradigms to better understand and simulate pain, particularly in controlled environments such as medical training systems. 

\begin{figure}[!t]
   \centering
    \includegraphics[scale = 0.6]{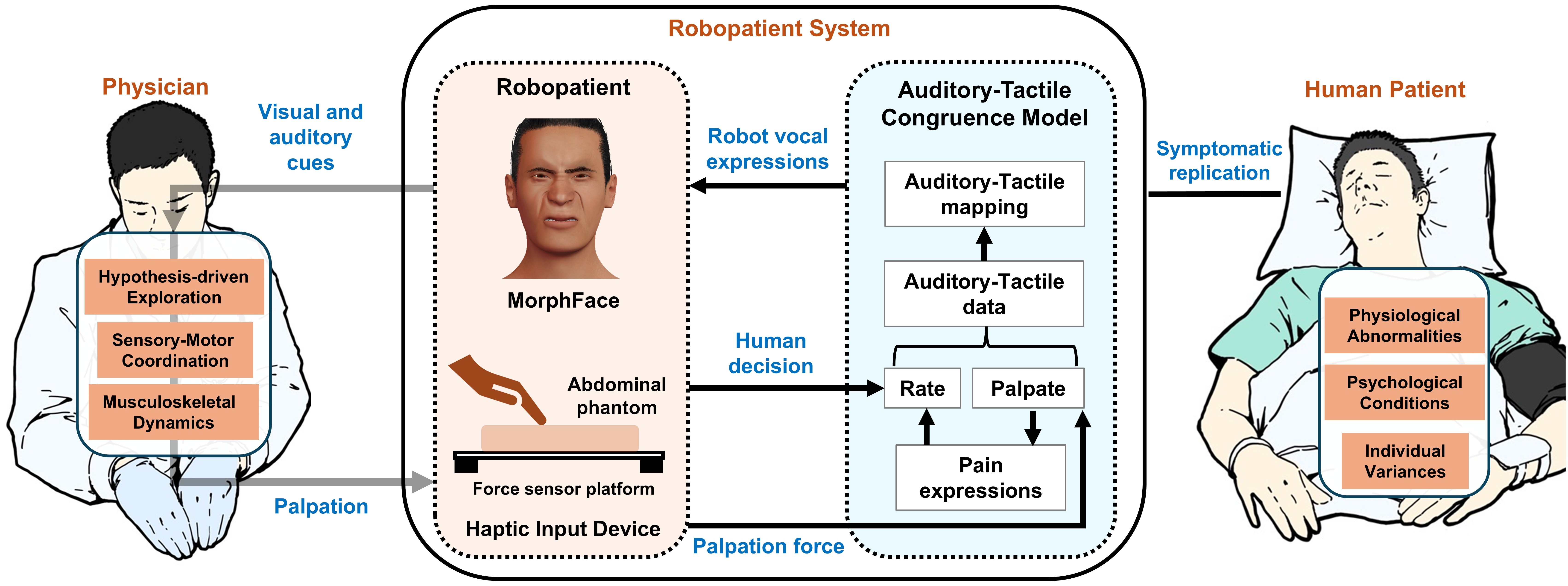}
    \caption{Overview of the conceptual model for studying auditory-tactile congruence in robopatients. The system comprises three main components: the user (ideally a physician), the robopatient, and the human, providing feedback to train the robopatient. The human patients under examination may exhibit various physiological abnormalities, psychological conditions, and individual characteristics, all of which influence the dimensionality of pain expression in response to palpation. The robopatient serves as an adaptive intermediary, learning appropriate pain responses from the user and aiming to replicate human vocal pain expressions (including facial pain expressions for realism) realistically. During palpation, the robot’s abdominal phantom captures the user’s haptic input, which is recorded and processed through the robopatient’s internal palpation-pain mapping model. This conceptual model is derived from data collected from participants of diverse backgrounds during previous studies \cite{lalitharatne2022face, protpagorn2023vocal, 11177232}, and these studies inform a universal method for mapping palpation to pain expressions, formulated as the ``auditory-tactile congruence model''. Then the robot outputs corresponding pain sounds and facial expressions through MorphFace \cite{lalitharatne2021morphface}. The user provides feedback on the robopatient’s generated pain responses, which were later refined by comparing palpation data with user feedback to map the relationship between palpation and the pain expressions it causes. This concept was adapted from \cite{lalitharatne2022face} and modified to fit the new paradigm.}
    \label{overview}
\end{figure}

Due to the risks associated with training on actual patients in nursing situations, simulation-based education (SBE) is used to train medical students. SBE enables a safe and controlled environment \cite{alinier2022simulation} for medical students to practice their physical examination, including abdominal palpation. Social science rises visibly above the level of common sense in contributing to our understanding of communication in the patient-physician relationship \cite{ charney1972patient}. Standardized patients (SPs) or professionally trained actors acting as patients, have been the highest fidelity simulators used in the medical field for decades \cite{costanza1999effectiveness}. SPs were capable of replicating facial expressions, vocal or audio expressions, both in the form of meaningful sentences or unstructured grunts, and variations of muscle stiffness during palpation training. Although this is effective, it is time-intensive to train and maintain their skills. Hence, we introduced robopatient \cite{lalitharatne2021morphface}, a medical robotic training simulator that helps ground the emotions associated with a patient and aid the diagnosis process. This further offers a platform for studying complex subjective psychophysiological phenomena, such as pain, by translating them into interpretable embodied expressions through robots. 

\vspace{3mm}

In this work, we introduce a novel approach for generating vocal pain expressions by integrating human feedback. This platform combines dynamic vocal pain sounds and facial expressions to simulate human-like pain responses corresponding to palpation forces. This concept, visualized in Figure \ref{overview}, comprises a force-sensitive abdominal phantom equipped with load cells and a controllable 3D hybrid face, MorphFace \cite{lalitharatne2021morphface}, capable of displaying facial and vocal pain expressions. An experimental study was conducted involving 7680 trials across 20  participants (10 males, 10 females) performing palpation tasks on a robopatient across set target force levels. The real-time auditory and visual feedback was generated by the robot, directly mapping the palpation to pain, and allowed participants to rate the perceived intensity of pain sounds. Selected properties of pain expressions were changed to find the contribution of gender (male and female sounds) and acoustic features (amplitude and pitch) of pain sounds towards perceived pain during palpation.  This systematic investigation into the relationship between properties of pain sounds and force levels offers a unique framework for exploring pain perception in medical training. 

\vspace{3mm}

The rest of the paper is organized as follows: The prior work and its limitations are discussed in section \ref{related_work_sec}. The implementation of the synthetic pain sounds, their selection, and the hardware system of the robopatient are discussed in section \ref{pain_generation_sec}. The experiment methods and protocol used to conduct the study with human participants are described in section \ref{experiments}. The results of the study are discussed in section \ref{Results_and_discussion} with the data curation and analysis techniques used. The post-study questionnaires completed by the participants are also analyzed and discussed in this section. The limitations of the current study, its implications, and the future directions of this work are also mentioned in this section. Finally, we conclude in section \ref{conclusion} by discussing the key findings of this work and its contributions that could lead to the development of robotic patients for medical training.

\section{Related Work}\label{related_work_sec}

Robots can be useful for studying complex psychophysiological phenomena, such as pain, by simulating human-like responses through biologically inspired models. The Brain-inspired Robot Pain Spiking Neural Network (BRP-SNN) that mimics the neural pathways involved in pain perception is an example of enabling robots to exhibit adaptive pain responses to physical stimuli \cite{feng2022brain}. Additionally, a care training assistant robot with 3D facial pain expressions can enhance caregiver training by providing realistic visual pain cues \cite{lee20213d}. These robotic systems allow for controlled, repeatable studies of pain mechanisms and improve training by offering consistent, interpretable simulations of human pain responses.

\vspace{3mm}

Computational pain recognition has two sides: clinical and social. The clinical aspect of pain includes the detection and assessment of the intensity of pain \cite{ cao2021can}. 
Humans vocalize different types of pain sounds in different contexts, such as vocal pain expressions during abdominal palpation and when someone touches a hot object. Hence, modeling pain sounds or vocal pain expressions is challenging \cite{lautenbacher2017phonetic, helmer2020crying}. In addition, pain perception and expressions vary depending on human backgrounds, such as gender \cite{rhudy2010there} and ethnicity \cite{racine2012systematic}. When infants experience a higher intensity of pain, they produce more irregular cries \cite{tiezzi2004determination} with higher amplitude \cite{lehr2007neonatal, fuller1995effect}, larger fundamental frequency variation \cite{porter1986neonatal, koutseff2018acoustic}, smaller amplitude variation, and for longer duration \cite{porter1986neonatal, johnston1987acoustical}. Hence, the computational pain recognition must account for both clinical assessment and social context, as vocal pain expressions vary widely based on the situation, individual background (e.g., gender, ethnicity), and intensity of pain. This variability makes modeling pain sounds complex and context-dependent. Facial pain signals can sometimes operate independently of voice generation, influenced by features such as fundamental phonation frequencies and intonation \cite{campanella2007integrating}. However, sound is more informative than visual signals in expressing emotions \cite{8441363}, making it a vital modality for simulating pain in computational assistive systems. By incorporating multimodal signals, including synthetic vocal expressions that mimic human pain, robotic systems can foster human-like empathy in contexts such as nursing and healthcare \cite{pepito2020intelligent}. Despite its inherently interdisciplinary nature spanning neuroscience, psychology, social contexts, and computer science, the study of pain and its synthetic representation has yet to fully take advantage of these connections. This highlights the need for innovative, cross-disciplinary approaches to better understand and simulate pain in training systems and assistive technologies.

 \vspace{3mm}
 
There exists a correlation between bio-signal parameters related to speech prosody and self-reported pain levels \cite{oshrat2016speech}. Therefore, non-linguistic information or ``how'' we speak could sometimes be more important than the linguistic information or ``what'' we say. In addition, gestures could be the most accurate indicator of pain in an unimodal emotion recognition system, followed by speech and facial expressions \cite{ castellano2008emotion}. Hence, gestures could be accurately read by a doctor during the palpation, while sound and facial expressions can be confusing. Robopatients, on the other hand can simulate these two confusing features in order to assist doctors in recognizing pain demonstrated by facial expressions and non-linguistic sounds. For this, understanding human pain is crucial.

\vspace{3mm}

Due to the complexities involved in modeling pain sounds in humans, only a handful of studies have been reported. A recent study on pain analyzed the acoustic characteristics that are responsible for variation in pain ratings in pain simulation by trained actors \cite{raine2019vocal}. This study identified that a range of voice fundamental frequency (F0, perceived as pitch), the amplitude of the vocalization, the degree of periodicity of the vocalization, and the proportion of the signal displaying non-linear phenomena all increased with the level of simulated pain intensity. Thus increasing the perceived pain by the listeners. Although these actors closely imitate infants and nonhuman mammals' pain vocalization, pain studies related to adult humans are scarce.

\vspace{3mm}

When robots recognize human pain signals, such as painful facial expressions, and display pain-related empathy, people often adjust their own biophysical signals to make their internal state more easily `read’. Pain make sense in robots in two different forms: pain detection and pain generation. Our study is on pain generation for a robopatient, which should cover diverse ethnicities, genders, and ages. Elicited expressions, which indicate expressions evoked by audiovisual media or real interpersonal interaction, have been generated and projected on the robopatient. Gender differences in the experience and processing of emotion contribute to differences in the modulation and perception of pain due to differences in the modulation and perception of pain \cite{ rhudy2005gender}. In addition, there have been robotic platforms that studied concepts underlying physical pain, such as SEATTLE \cite{asada2019artificial}, which resembles an artificial ``pain nervous system''. However, the multidimensional nature of pain, which extends beyond the nervous system, remains poorly understood.

\vspace{3mm}

Considering the above aspects of palpation and pain expression, we tailored our study to gain insights from people to develop a robopatient that could replicate vocal and facial expressions related to pain during palpation. We identified that there are no subsequent models that could map voice and facial expressions simultaneously to pain during palpation. To develop such a model, we hypothesized that data-driven methods can model the relationship between the palpation behaviour and vocal expressions, with the help of reverse correlation and user studies. A similar data-driven approach was used in prior work, which was based on audio-tactile cross-modal congruence for surface texture classification \cite{liu2022texture}. To this end, we conducted a study where the participants could select appropriate pain expressions in terms of voice and facial expressions to represent different types of pain created by palpation. We used the force applied during the palpation as a parameter to demonstrate the level of pain a patient will experience during palpation. In the end, we present the haptic-auditory-visual scenario of a patient in our model during palpation. 

\vspace{3mm}

A controllable and customisable physical robopatient with facial and vocal pain expressions for abdominal palpation training has been developed in \cite{protpagorn2023vocal}. The robopatient setup consists of a robotic face called MorphFace \cite{lalitharatne2021morphface} and a force sensor platform. This robotic platform has been used to render real-time pain facial expressions and vocal pain expressions based on the user's palpation force and position on a silicone abdominal phantom. A simple yet effective threshold-based algorithm that was built upon our previous study \cite{protpagorn2022vocal} for generating the vocal pain expressions based on the estimated pain intensity upon palpation, has been implemented on this robopatient.

\section{Synthetic Pain Generation in Robopatient}\label{pain_generation_sec}

\begin{figure}[h!]
   \centering
    \includegraphics[scale = 0.55]{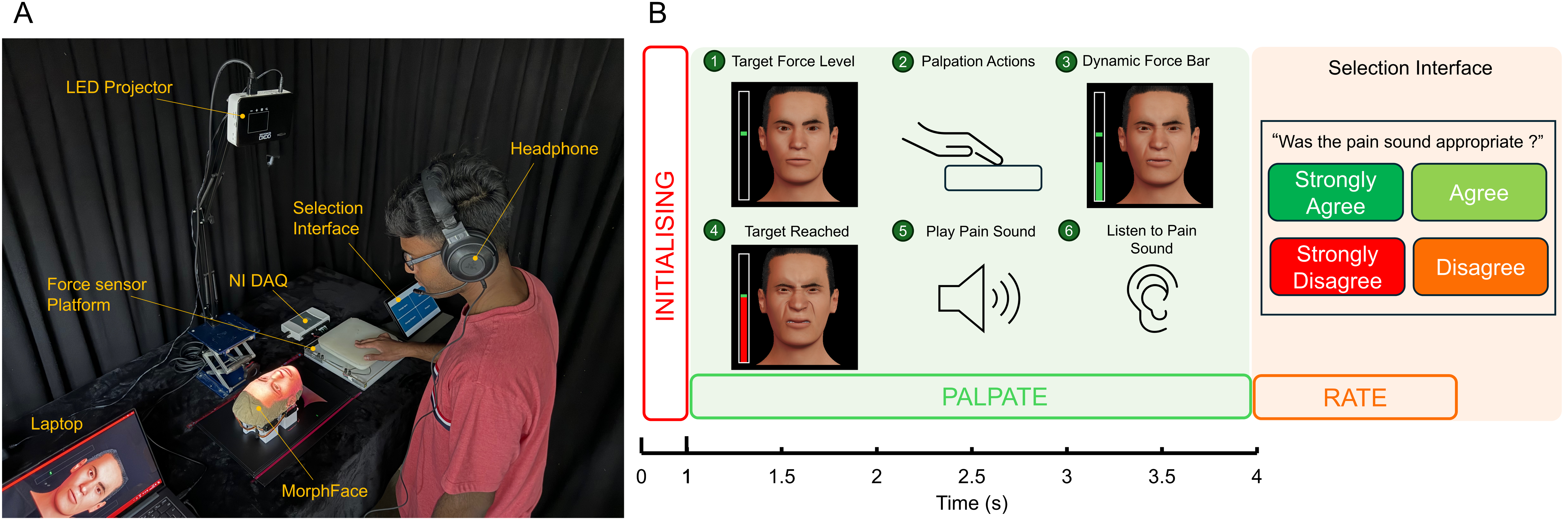}
    \caption{ Physical robot and the experiment procedure A) Robopatient setup. A scenario where the participant is palpating the robopatient is shown. B) Data flow within the robopatient setup during the experiment. The interaction between the robot and the user starts as the user palpates the abdominal phantom (Steps 1, 2). The force the user applies has to reach a certain level indicated by a progress bar next to the robot's face (Step 3). The progress bar is initially green and turns red when the required minimum amount of force is applied.  Based on the palpation force, the robot generates facial expressions and sounds independently of each other (Step 4). Then the user listens to the pain sounds generated by the robot for the palpation force (Steps 5, 6) and provides feedback on whether they agree with the robot’s mapping of pain sounds to palpation.}
    \label{trial_sequence_final}
\end{figure}

Facial and auditory pain expressions (pain sounds) were synthesized using robopatient. This robopatient consists of an abdominal phantom to mimic the abdomen of a patient and MorphFace to synthesize facial expressions. Robopatient and experimental procedure are shown in Figure \ref{trial_sequence_final}. 

\subsection{Selection of Pain Sounds }

For the experiment, three different pain sounds were selected per gender. This is to observe whether participants perceived certain pain sounds better than others. The pain sounds that were deemed appropriate were selected from the pilot trials. The pain sounds were selected from an online sound effects platform (\href{https://www.soundsnap.com/}{\color{blue}Soundsnap}). Male and female pain sounds were played to observe whether participants of a certain gender (male or female) may perceive the corresponding pain sounds better or worse. The \textit{audioread} function in MATLAB 2024a (Mathworks, Inc.) was used to import audio files and individually obtain the amplitude and frequency (pitch) of the audio. The \textit{sound} function in MATLAB was used to play the pain sounds. The male and female sounds selected for the experiment mentioned in section 4 are provided \href{https://github.com/nsaitarun-git/Submission_Codes/tree/main/Pain_Sounds}{\color{blue}here} with their designated pain sound numbers (i.e., 1, 2, and 3) appended to the end of their filenames.

\subsection{Abdominal Phantom}

A force sensor platform and silicon phantom replicated a human's abdomen as shown in Figure \ref{overview}. The force sensor platform was developed to measure the palpation force during user palpation. The force sensor platform was made of four single-axis (20 \textit{kg}) load cells (Haljia, Inc.). The load cells were mounted at the four corners of a generic square shape (250 $\times$ 250 \textit{mm}) acrylic platform. The distances between two adjacent load cells were 246 \textit{mm}. A National Instruments DAQ (USB-6211) with MATLAB was used to acquire the signals from the load cells at a sampling rate of 1000 \textit{Hz}. The force value of each load cell was used to calculate the total palpation force ($F_{tp}$). To minimize noise in the total palpation force, values below 1N were neglected based on observations from pilot experiments.

\begin{equation}
  F_{tp} = \sum_{j=1}^{4} F_{j}
\end{equation}

where $F_{j}$ is the force reading from the $jth$ load cell mounted on the four corners of the force sensor platform.

\vspace{3mm}

A square-shaped (200 $\times$ 200 $\times$ 30 \textit{mm}) silicone block was fabricated with Ecoflex 00-30 (Smooth-On, Inc.). We placed this silicone block on the force sensor platform. It was used as the abdominal phantom of the simulated human.

\subsection{MorphFace}
MorphFace is a controllable 3D physical-virtual hybrid face that can represent facial expressions of pain displayed on six different ethnicity-gender face identities: female and male faces of White, Black, and Asian ethnicity. The pain facial expressions are realized by 4 facial action units: AU4 (Brow Lowerer, AU7 (Lid Tightener), AU9 (Nose Wrinkler), and AU10 (Upper Lip Raiser), with the pain intensity $(PI)$ ranging from 0 (no pain) to 100 (maximum pain). More details of MorphFace can be found in our recent work \cite{lalitharatne2021morphface}. We included only the white male and white female face identities for this study. The servo motors attached to different parts of the face used to realise action units, were not actuated for this study. Hence the physical face was static while the projected facial expression was dynamic over time, based on the pain intensity.




\section{Experiments} \label{experiments}

We asked 20 healthy participants (10 males and 10 females) aged between 21 and 50 years (mean age 30.05 years, SD 9.5 years) to take part in this study. Students and faculty with no prior experience in pain to palpation mapping and no history of sensorimotor impairments were chosen to participate in this experiment. Refer to Figure \ref{trial_sequence_final}A for the experiment setup. We conducted our study with an equal number of participants from each gender to keep gender balance and study gender-related trends as well \cite{rhudy2005gender}.

\vspace{3mm}

The experiment consists of one task. The participant must palpate the abdominal phantom until they reach a target force. Four levels of target forces were presented to each participant to ensure they applied a diverse range of forces when palpating. A pain sound was played when the applied force was equal to or greater than the target force. We refer to the set target level as the \emph{Force Threshold} (5/10/15/20 N) and to the measured total palpation force as $F_{tp}$ (in N). The participant must then rate whether they `strongly agree' (a), `agree' (b), `strongly disagree' (c), or `disagree' (d) through the selection interface. We labeled each choice with a letter (a-d) to make data management more intuitive. We aimed to explore the relationship between palpation force and the creation of a palpation-to-pain sound mapping across a range of forces. Therefore, the target force was used as a trigger or control variable. In a previous attempt to map pain responses to palpation force \cite{11177232}, we observed complex relationships between physical palpation and auditory pain expressions that contradicted our initial hypothesis. Using proximal policy optimization (PPO), the study learned individual preferences that linked specific forces to pain expressions, indicating that personalized mappings were more effective than a generalized model. However, the study involved only a small number of participants, motivating the present expanded investigation into the relationship between physical and expressive aspects of pain.

\vspace{3mm}

For the experiments, four parameters were changed between the trials. The target force, pain sound, pain sound amplitude, and pain sound pitch. Four different values were used for the target forces, pain sound amplitude, and pain sound pitch. We determined these values would be appropriate from pilot trials. We only chose three individual pain sounds for this scoping study, as the combination of pitches and amplitudes with relation to an individual pain sound can be tremendously large and is very difficult to study. Furthermore, we only chose to change these four significant parameters during the trials, as changing no more than four variables simultaneously ensures it is within the human capacity to reliably interpret and respond during affective or perceptual tasks \cite{breazeal2003emotion},\cite{halford2005many}. 

\begin{equation}
    Amplitude = [1,1/3,1/9,1/27]^T
\end{equation}

\begin{equation}
    Pitch = [0.7,0.9,1.1,1.3]^T
\end{equation}

\begin{equation}
    Force \ Level = [5,10,15,20]^T
\end{equation}

\begin{equation}
    Pain \ Sound = [1,2,3]^T
\end{equation}

192 trials were conducted for each gender ( male and female). The total number of trials per participant was 384. The experiment was divided into two parts: robopatient displaying male and female expressions at a time. In each part, the facial expressions displayed on MorphFace and pain sounds played were selected based on a particular gender. This was to minimize cognitive load, which can be introduced by randomly changing the facial and vocal expressions of the robopatient to the opposite gender during the trials. A matrix of all possible combinations of amplitude, pitch, force levels, and pain sounds was created in MATLAB. A combination of the four parameters was selected once at random from the matrix for each trial. None of the combinations were repeated for that gender. 

\vspace{3mm}

Each trial lasted 3 seconds, and initializing the next trial took 1 second. There was no time limit for the participants to make their decision. During the 3-second trial, the total palpation force was acquired through the DAQ. When a trial began, the participant was presented with a graphical interface with a green colored marker (target force) and a face image on MorphFace. A dynamic green bar appeared, adjusting its length in real time according to the force applied during palpation of the silicone block. 
Once the participant reached the target force, the pain sound was played through headphones (Razer Kraken X Lite), and the dynamic green bar became static and turned red. The program then waited for the participant to rate the pain sound. We asked the participants to choose their response based on how appropriate the pain sound they heard was in relation to the palpation force (i.e., strongly agree, agree, strongly disagree, and disagree). The program then moved on to the next trial. Every 64 trials, the participant was given a 90-second break. At the end of the 192 trials, a 5 minutes break was given before starting the trials for the next gender. 

\vspace{3mm}

The participants were allowed to familiarize themselves with the system. To do this, six familiarization trials were conducted for each gender. None of the parameters were randomized during these trials. The pain sounds were played without alteration. The force levels changed in ascending order for each trial. The force levels for the 5th and 6th trials were 5 N and 10 N. Counterbalancing was introduced to this study by changing the order in which the participants began. For example, if a participant first started the experiment with male pain expressions, the next participant started the experiment with female pain expressions. This was to reduce bias, and the experiment took approximately one hour to complete for each participant. This experimental process is shown in Figure \ref{trial_sequence_final}B. 

\vspace{3mm}

The participants spent well over 37 minutes on the main experiments, which include the four 90-second breaks and the 5-minute break. They spent more than 48 seconds on the familiarization trials. The exact experiment time of each participant depends on their decision time. However, none of the participants took more than an hour to complete the study, which included their briefing.

\vspace{3mm}

We used the $(PI)$ values to map the length of the green bar and the faces projected on MorphFace. We used a force range of 20 N (maximum target force) to map $F_{tp}$ to $PI$. We used a scaling factor ($\beta=5$) and $F_{tp}^{filtered}$, which was the filtered output (using a moving average filter of window size 20) of $F_{tp}$ to determine the values of $PI$. The facial expressions projected on MorphFace and the length of the dynamic green bar changed between 0-100 based on the value of $PI$.

\begin{equation}
    PI = \beta F_{tp}^{filtered}
\end{equation}


\vspace{3mm}
Though the main focus of this scoping study is exploring audio-tactile congruence, we believe the integration of facial expressions is necessary alongside the vocal pain sounds during the experiments. Facial pain expressions would provide context and help participants perceive the pain sounds effectively; otherwise, the sounds on their own can be odd and confusing. Despite the integration of facial pain expressions, we believe they would not affect the nature or findings of this study, as the responses of the participants would be dependent on the pain sounds.

\section{Results and Discussion}\label{Results_and_discussion}

\subsection{Dataset}
 The data of each participant comprised multiple features, capturing the force information from palpation, corresponding pain sound, amplitude, and pitch generated by the robot and participant's choice (agree-disagree). A section of this dataset is shown in Table \ref{T1}. In this section, we present the results of our analysis, including dimensionality reduction, clustering, association rule mining, and the interpretation of key findings from these tests.



\begin{table}[!b]
\centering
\caption{Dataset}
\label{T1}

{\small 
\begin{tabular}{p{2cm} p{2cm} p{3cm} p{2cm} p{2cm} p{2cm}}
\hline
\textbf{Force Threshold} & \textbf{Pain sound} & \textbf{Gender of pain expression} & \textbf{Pitch} & \textbf{Amplitude} & \textbf{Choice} \\ \hline
5 & 2 & F & 0.9 & 0.33333 & d \\
20 & 2 & F & 0.7 & 1 & b \\
15 & 3 & M & 0.9 & 0.11111 & d \\
10 & 3 & M & 1.1 & 0.11111 & b \\ \hline
\end{tabular}
}

\begin{tablenotes}
\small
\item *Pain sounds are indicated by numbers (1–3), representing three distinct audio tracks. ``F" and ``M" indicate female and male genders, respectively. For choices, a–d represent ``strongly agree" to ``disagree" (refer to section \ref{experiments} for choice labelling).
\end{tablenotes}
\end{table}

\subsection{Results}

\begin{figure}[h!]
    \centering
    \includegraphics[scale = 0.8]{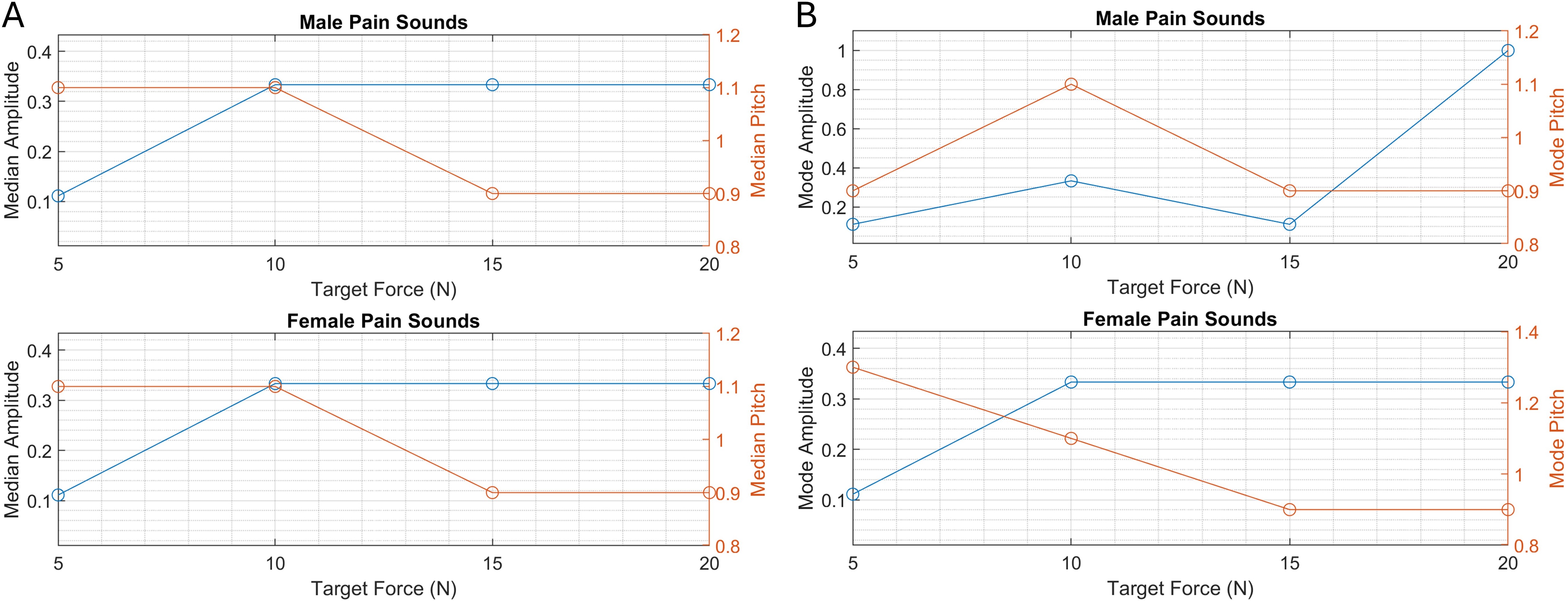}
    \caption{A) Median values of all participants' responses (`strongly agree'/`agree') for the amplitude (primary y axis- shown in black) and pitch (secondary y axis- shown in orange) of male and female pain sounds. B)  Modes of all participants' responses for the amplitude (primary y axis- shown in black) and pitch (secondary y axis- shown in orange) of male and female pain sounds. }
    \label{fig1}
\end{figure}

The study's primary aim was to explore the relationship between the acoustic properties of sound and the applied palpation force. The median and mode amplitude and pitch values for each target force are presented in Figure \ref{fig1}A and B. This analysis was conducted for trials where the participants chose `strongly agree' and `agree'. The observed results show that the preferred median amplitude of male pain sounds is at a minimum at the beginning, and reaches a maximum after 10 N and remains constant beyond that. The preferred median pitch remains constant until 10 N and falls to a minimum beyond that. Similarly, for female pain sounds, we observe that the preferred mode amplitude of pain sounds is at a minimum at the beginning, and reaches a maximum after 10 N and remains constant beyond that. The preferred mode pitch gradually decreases until 15 N and remains at a minimum beyond that. However, a different trend can be observed for mode amplitude and pitch values for male pain sounds. Both amplitude and pitch start at a minimum when the target force is 5 N. They reach a maximum at 10 N and fall when the target force is 15 N. Pitch remains constant beyond that, and amplitude increases to 1. Hence, this trend can be observed with both male and female pain sounds. 

\vspace{3mm}

These findings suggest that subjective and perceptual factors, as well as the unique acoustic properties of male and female pain sounds, play a significant role in shaping participants' preferences. Further research could investigate how sound properties are processed at higher force levels and whether these preferences are influenced by cultural, emotional, or physiological expectations of pain representation.

However, intuitively, pain induced by an applied force would be expected to increase with both sound pitch and amplitude. Several factors may account for the discrepancy:

\begin{enumerate}
       \item Perceptual saturation point\\
    People may experience a perceptual saturation point, where increasing the force level no longer significantly enhances their perception of pain intensity through sound. After reaching a certain amplitude, additional increases in force may not require correspondingly higher sound intensities for recognition \cite{melara1990interaction}.
    \item Cognitive Expectations and Adaptation\\
Participants might cognitively align their expectations of pain intensity with what feels reasonable or familiar. They might interpret excessively loud or intense sounds as unrealistic for high force levels, leading to a plateau in their preference for amplitude \cite{khera2021cognition}.
\item Gender-Related Vocal Characteristics\\
Male and female pain sounds may have inherent differences in timbre, frequency range, or emotional interpretation. These differences could subtly influence the force level at which participants perceive the sound as matching their expectations of intensity, creating slightly different thresholds \cite{keogh2014gender}.
\item Contextual or Experimental Bias\\
Randomized presentation of amplitudes and pitches might have introduced inconsistencies in how participants mapped sounds to forces. This randomness may have made participants favor certain sound properties irrespective of increasing force \cite{sussex2015different}.
\item Emotional Overlap\\
At higher force levels, participants might not perceive additional increases in amplitude as more reflective of ``pain'' but instead as overly dramatic or even distressing, which could cap their preference for sound amplitude \cite{cuadrado2020arousing}.
\end{enumerate}

These features have been observed in previous pain-related studies, but larger-scale experiments will be needed to confirm these behaviors in robotic patients.

\vspace{3mm}

\begin{figure}[h!]
    \centering
    \includegraphics[scale = 0.7]{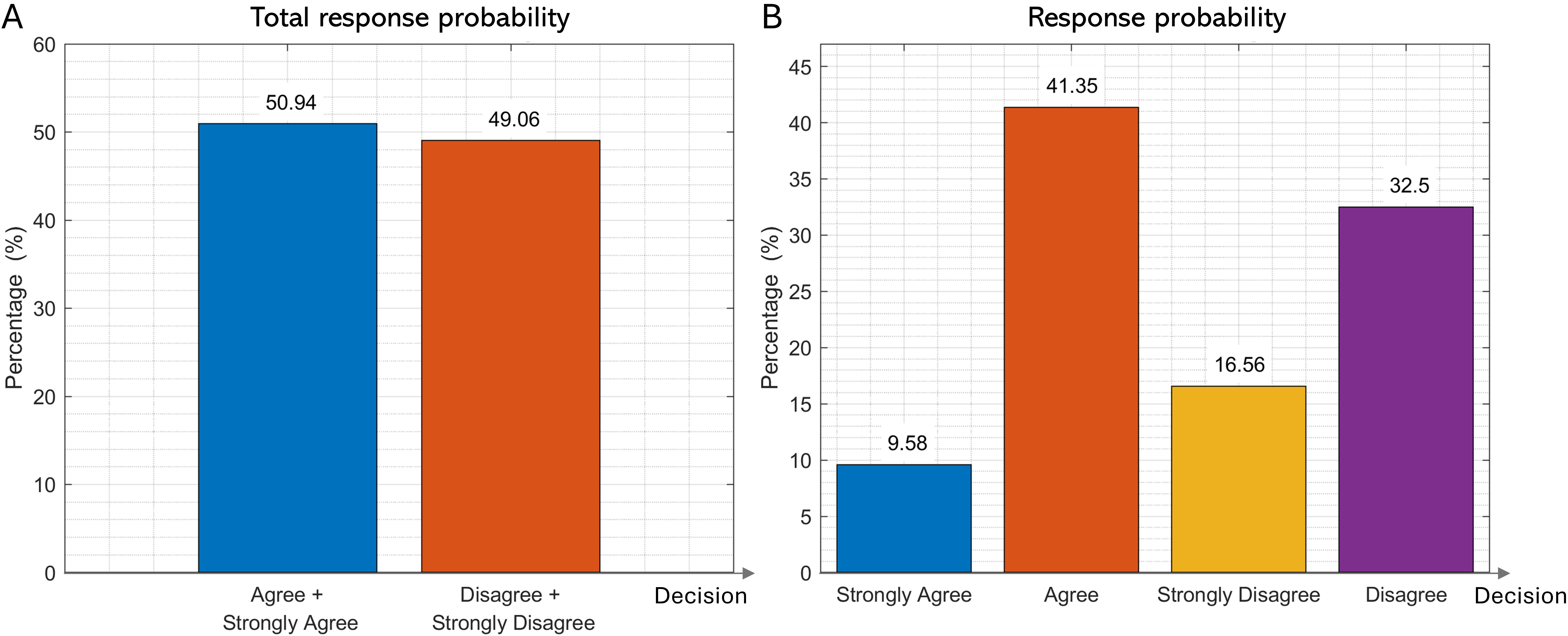}
    \caption{A) The responses given by the participants (Agree/Strongly agree and Disagree/Strongly disagree) for all the trials throughout the experiment are presented as percentages. B) The response probability of each response is shown here.}
    \label{Barchart}
\end{figure}

To gain a more discrete understanding of whether participants agreed or disagreed with the pain sound combinations we used in the experiments, we combined the strongly agree/ agree and strongly disagree/ disagree to create two distinct groups. This is done according to the common response collapsing scheme \cite{van2020criteria}, where we collapsed the four response options into two (agree/disagree) for binary analysis. The response percentages for agree/ strongly agree and disagree/ strongly disagree across all trials are shown in Figure \ref{Barchart}A. From the given responses—agree/ strongly agree: 50.9375\%, and disagree/strongly disagree: 49.0625\%, near equal opinions could be observed for both cases. Hence, the responses are nearly evenly split between agreement and disagreement. The close split suggests that the task or question may be subjective, leading to varied interpretations or individual differences in how participants perceive and match pain sounds to force levels. This is more clearly seen in Figure \ref{Barchart}B, where a large variation can be observed between the split of strongly agree and agree compared to strongly disagree and disagree. Furthermore, from these results, we assume that an adequate number of choices of pain sounds were covered during the experiment.

\begin{figure}[h!]
    \centering
    \includegraphics[scale = 0.85]{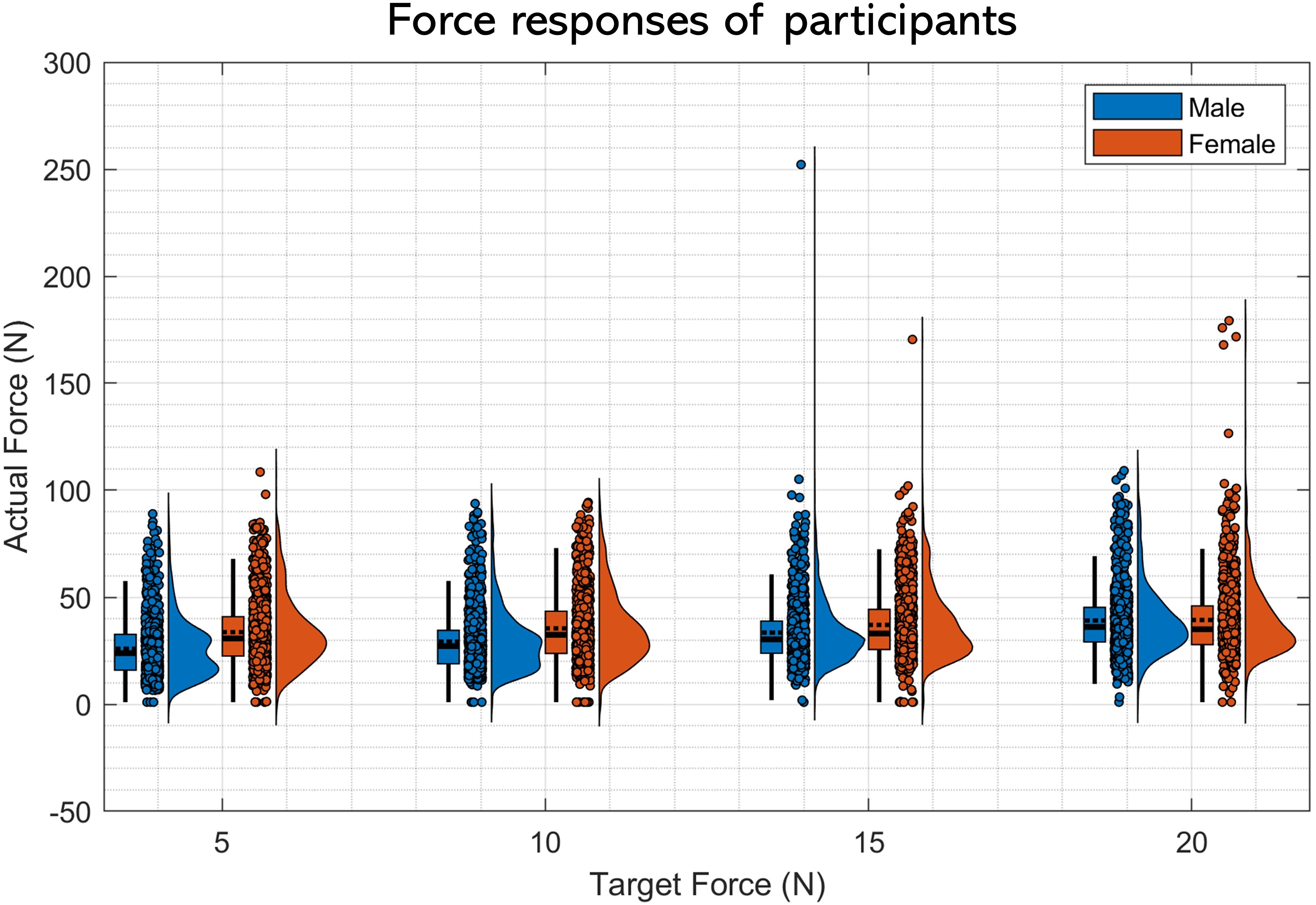}
    \caption{Violin plot visualizing the distribution of actual force responses across different target forces. The width of each violin reflects the density of actual forces, with the central line indicating the median and the shaded area showing the inter-quartile range (IQR). This visual highlights differences in actual force distributions among the four target force categories. The plots represent the force responses for male and female participants for each target force, respectively (refer to legend). The dashed black lines represent the mean actual force of each target force.}
    \label{fig2}
\end{figure}

\vspace{3mm}
When the actual force distribution was plotted against each target force level, a slight increase in actual force was observed as the target force levels increased. This trend is illustrated in Figures \ref{fig2} A and B. Taking into account the peak force distribution and mean values, we observed a small increase in the forces exerted by the female participants. We saw this pattern in another study \cite{11177232} that we conducted within the same context as well. Hence, this could potentially suggest differences in force perception, motor control, or task approach between genders. These variations may highlight the need to account for individual or group differences in force-related tasks, as perception and motor strategies may differ. A larger sample size of participants in future work may further reinforce these conclusions. A detailed analysis of gender-based differences in responding to diverse robotic pain expressions is provided in \cite{11177232}. In brief, variations in initial responses can stem from physiological and psychosocial factors, consistent with prior work showing that gendered vocal cues shape perceived urgency, pain tolerance, and response behavior. The work in \cite{11177232} further shows that adaptive expressions and real-time human feedback can address this issue.

\begin{figure}[h!]
    \centering
    \includegraphics[scale = 0.88]{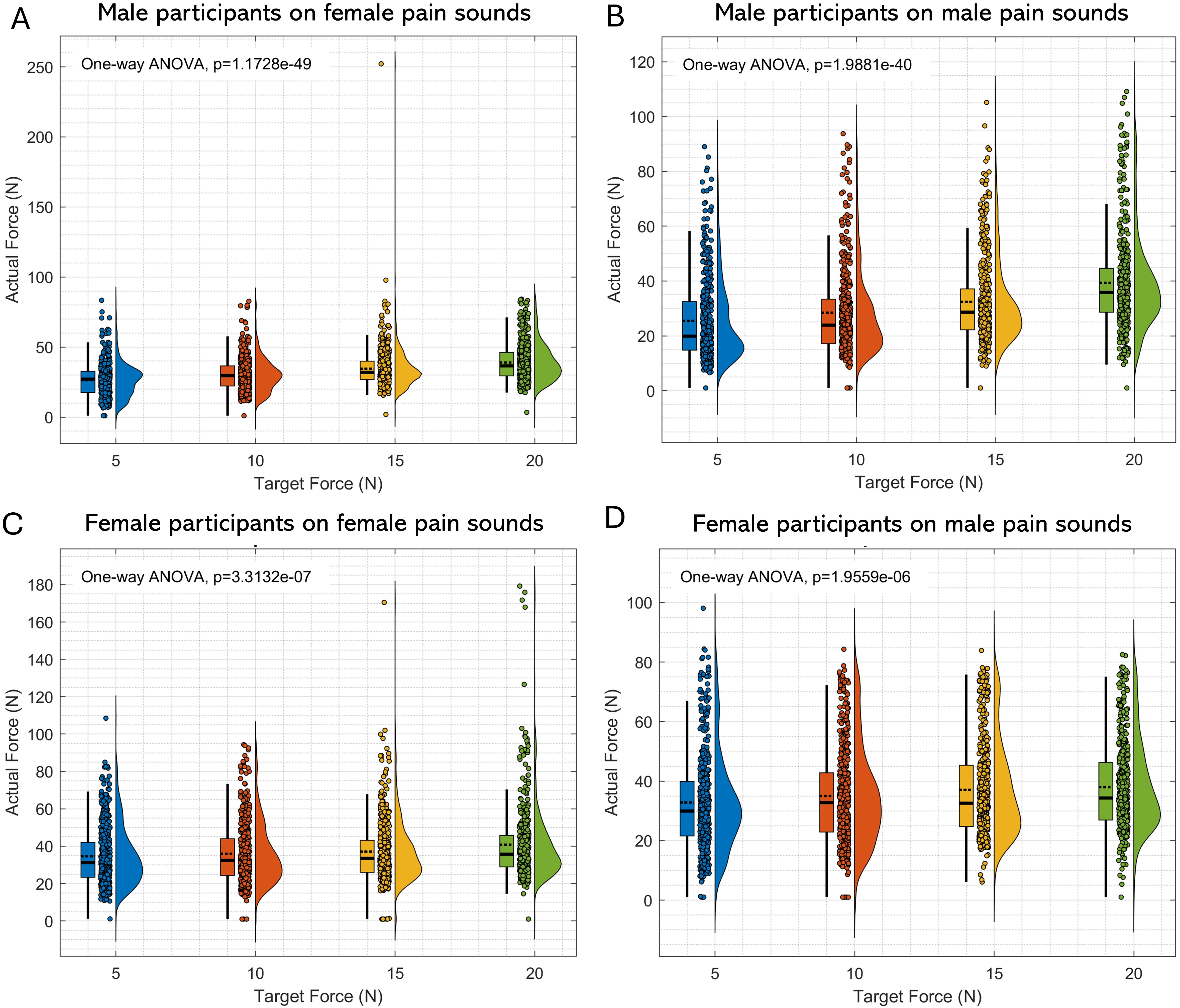}
    \caption{Violin plot visualising the distribution of actual force responses across different target forces. The width of each violin reflects the density of actual forces, with the central line indicating the median and the shaded area showing the interquartile range. This visual highlights differences in actual force distributions among the four target force categories. The plots represent the force responses of A) male participants on female pain sounds, B) male participants on male pain sounds, C) female participants on female pain sounds, and D) female participants on male pain sounds, respectively. The dashed black lines represent the mean actual force of each target force.}
    \label{fig3}
\end{figure}

\vspace{3mm}
A similar trend was observed in each gender's response to pain sounds of their own gender, as shown in Figure \ref{fig3}A-D. However, when examining responses to pain sounds of the opposite gender, the scatter increased significantly, with the data spread over a wider range in the violin plot. This may be attributed to differences in familiarity or emotional connection, as participants might find it harder to relate to or interpret the pain sounds of the opposite gender. Additionally, cognitive biases or emotional factors could influence how pain sounds are processed, leading to less consistent responses. These findings underscore the importance of accounting for gender-based perceptual differences in studies involving pain sounds, as they can significantly affect the interpretation of results and the design of such experiments.

\begin{figure}[h!]
    \centering
    \includegraphics[scale = 0.8]{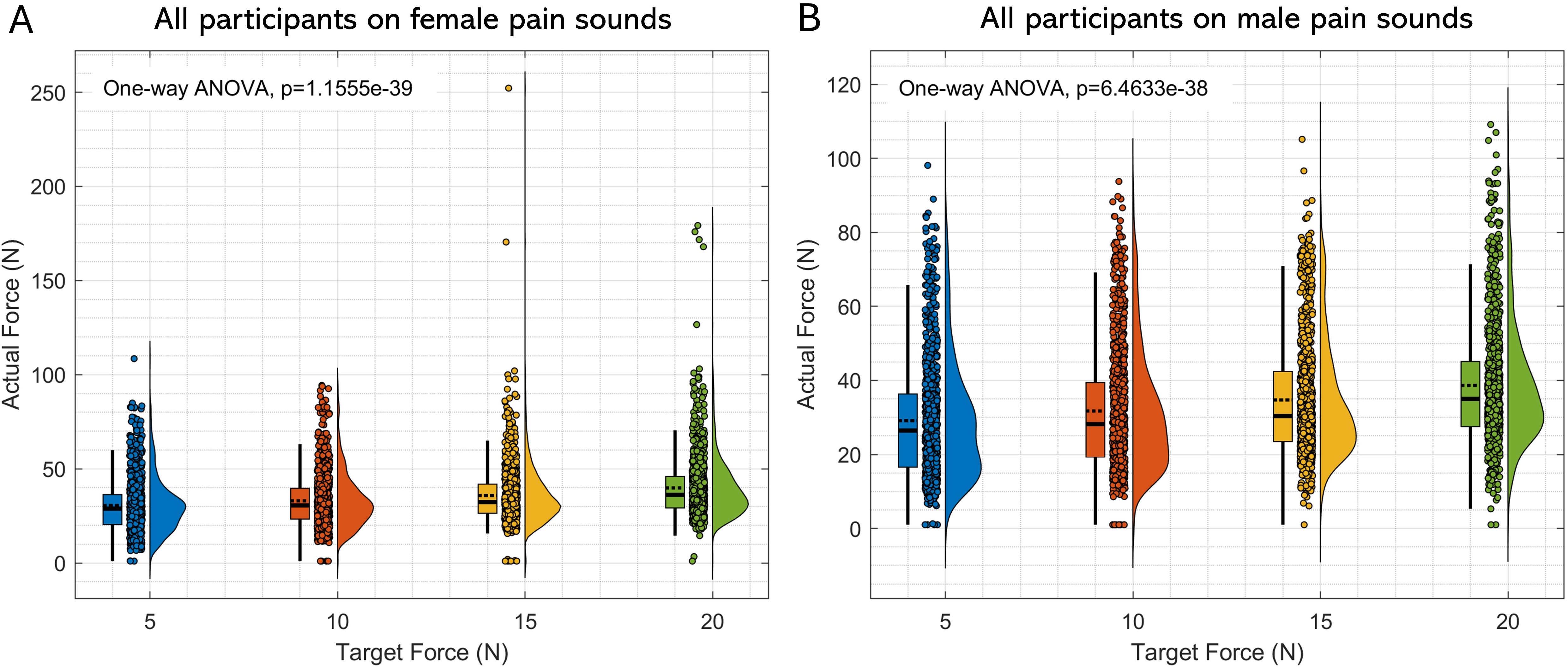}
    \caption{Violin plot visualising the distribution of actual force responses across different target forces. The width of each violin reflects the density of actual forces, with the central line indicating the median and the shaded area showing the interquartile range. This visual highlights differences in actual force distributions among the four target force categories. The plots represent the force responses of all participants, despite gender, on A) female pain sounds and B) male pain sounds, respectively. The dashed black lines represent the mean actual force of each target force.}
    \label{fig4}
\end{figure}

\vspace{3mm}
Interestingly, a similarly scattered pattern was observed in the overall responses to male pain sounds, as shown in Figure \ref{fig4}. Despite the variability, the distribution peaks consistently followed the same trend: a subtle increase in actual force with rising target force levels. The scattered response pattern for male pain sounds, combined with consistent peaks showing a subtle increase in actual force with higher target levels, suggests underlying consistency amid variability. Given the variability across the participants, our findings potentially indicate a relationship between applied force and perceived pain sounds; further analysis is required to confirm this. This potentially highlights the need for a task design that clearly elicits responses aligned with target force levels, even when sound characteristics introduce variability.

\vspace{3mm}

It should be noted that in Figures \ref{fig2},\ref{fig3}, and \ref{fig4}, participants apply forces that are higher and lower than the target force levels. We believe that participants can subjectively determine the necessary amount of force they would need to apply within the time limit to reach the target force. This could result in the maximum palpation force being higher or lower than the target force. Hence, some of the trials may result in the maximum palpation force being below the target force. While the actual forces differed from the target forces in many cases, we believe that the participants attempted to palpate in a way that matched the target force. Accordingly, as depicted in Figure \ref{fig1}, the results show a causal relationship in which amplitude increases with higher target forces, while pitch decreases. However, more trials with a larger number of participants can improve the insight into these findings.

\begin{figure}[!b]
    \centering
\includegraphics[width=0.95\linewidth]{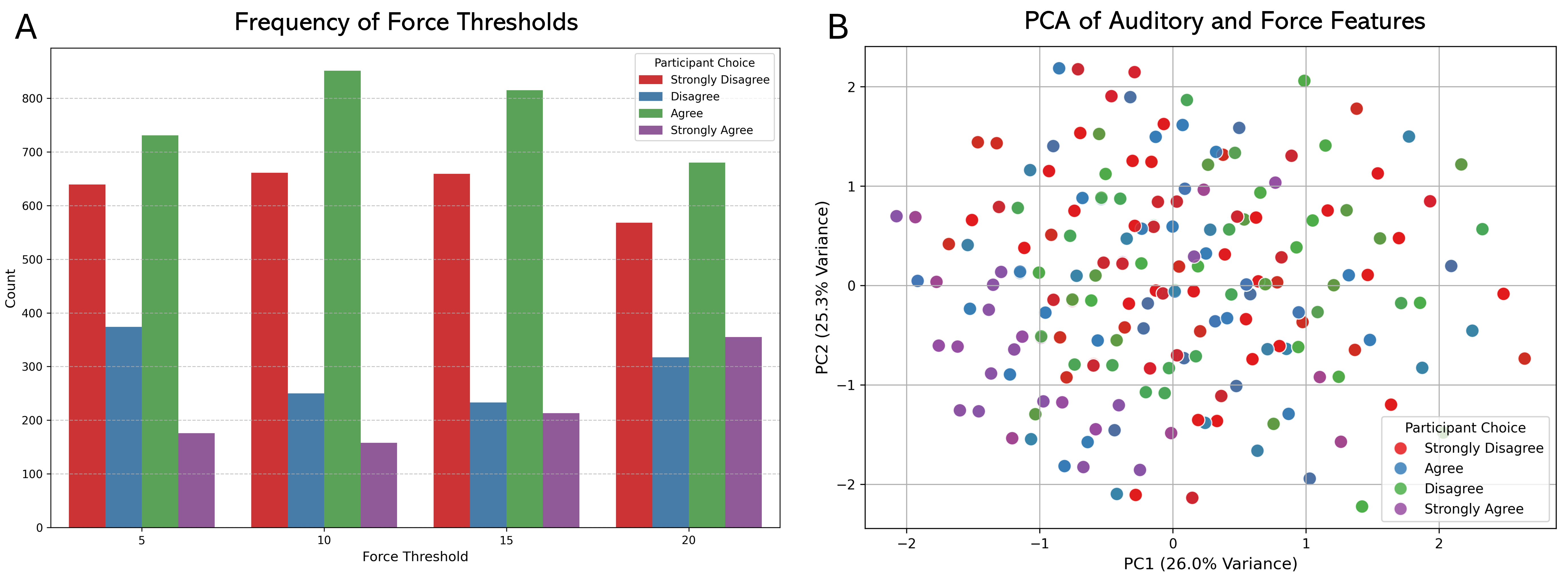}
    \caption{A) Frequency of participants' choices plotted against force thresholds. B) Exploratory PCA (visualization only; PCs from z-scored features) of auditory and force features. Both plots are colored by the participant’s choice.}
    \label{fig:ForceFreq}
\end{figure}

\subsection{Further analysis upon human feedback}

We analyze four predictors: \emph{Amplitude} (unitless loudness scaling), \emph{Pitch} (ratio levels), \emph{PainSound identity} (categorical), and \emph{Force Threshold} (the discrete target condition: 5/10/15/20 N). The total palpation force measured from the load cells is denoted $F_{tp}$ (in N). When principal components (PC1/PC2) are referenced, they are computed from standardized features (z-scores) and used for \emph{visualization only}; all statistical inference is performed in the native feature space (Amplitude, Pitch, PainSound, Force Threshold). One-way ANOVA $F$ statistics are defined as $F=\mathrm{MS}_{\text{between}}/\mathrm{MS}_{\text{within}}$, and post-hoc pairwise contrasts use Tukey’s HSD (based on the studentized-range distribution).

\vspace{3mm}

This section presents a data analysis of how auditory parameters (Amplitude, pitch, pain sound) and participant input (Force threshold) influenced perceived congruence of auditory expressions with tactile interaction. This analysis aimed to examine which features shaped participants' decisions (agree or disagree) with the sound-palpation force combinations in the context of palpation training. 

\vspace{3mm}

 Figure~\ref{fig:ForceFreq}A illustrates the distribution of force threshold values across participants' choices. It can be observed from the figure that Choice `b' (agree) is the most frequent response across all force levels, and Choice `a' (strongly agree) becomes more prominent at higher thresholds. The most likely reason for this behavior was increased perceived confidence or certainty by participants as the force threshold increases. At higher force levels, stimuli may become more easily perceptible or unambiguous, which leads participants to respond with greater agreement or stronger affirmation (i.e., choosing `a’ – strongly agree) instead of just `b’ – agree. This is further reinforced by `d' (strongly disagree) being the second frequently selected response for every force threshold. This consistent presence of strong disagreement suggests a subgroup of participants who perceived the stimuli as insufficient or ambiguous at lower thresholds, with frequency remaining stable from 5–15 N before decreasing at 20 N. Including both trends provides a broader picture of the response distribution and aligns with typical psychophysical patterns where stimulus strength influences both certainty and polarity of responses. This aligns with typical psychophysical response patterns where stronger stimuli yield more confident responses \cite{yoshida2013uncertainty, coghill2010individual, mcdaniel2023psychophysics}.

\begin{table}[!b]
\centering
\caption{ANOVA results and post-hoc (Tukey HSD) significance tests for each feature across participant Choice responses.}
\begin{tabular}{lccc}
\toprule
\textbf{Feature} & \textbf{F-statistic} & \textbf{p-value} & \textbf{Significant Pairwise Differences} \\
\midrule
Amplitude       & 29.65 & $<$0.001 & b $\ne$ d, b $\ne$ c, d $\ne$ a \\
Pitch           & 3.92  & 0.008    & b $\ne$ c, c $\ne$ d \\
Force Threshold  & 30.24 & $<$0.001 & a $\ne$ b/c/d \\
Pain Sound       & 0.44  & 0.722    & None \\
\bottomrule
\end{tabular}
\label{tab:anova}
\end{table}

\vspace{3mm}
We then performed one-way ANOVA tests to assess whether the amplitude, pitch, force threshold, and pain sound varied significantly across participant choices. Table~\ref{tab:anova} summarizes these results. This analysis revealed significant effects for amplitude (F = 29.65, p $<$ 0.001), pitch (F = 3.92, p = 0.008), and force threshold (F = 30.24, p $<$ 0.001), while pain sound did not differ significantly across participant responses (F = 0.44, p = 0.722). Post-hoc Tukey HSD tests further clarified these effects: for amplitude, significant differences were observed between `b' (agree) and both `c' (disagree) and `d' (strongly disagree), as well as between `d' and `a' (strongly agree). This indicates that louder sounds were more likely to elicit agreement, while quieter or mismatched amplitudes led to disagreement. In the case of pitch, the significant contrasts were between `b' and `c', and between `c' and `d', suggesting that pitch plays a more nuanced role in participant perception. For force thresholds, the most notable finding was that `a' responses differed significantly from all other choices, implying that participants who strongly agreed tended to apply greater force. Overall, these results suggest that both auditory intensity and tactile input (force) strongly influenced perceived congruence, while speaker identity (pain sound) has little perceptual impact.

\vspace{3mm}

To visualize the relationship between these features, we inspected a PCA projection of the standardized predictors (Figure \ref{fig:ForceFreq}B). The projection illustrates that data points for agreement (Choice `b') cluster densely, whereas strong agreement (`a') and strong disagreement (`d') diverge along the primary components. This visual structure complements the ANOVA results (Table \ref{tab:anova}), which indicate that \textit{Force Threshold} and \textit{Amplitude} are the statistically strongest drivers of participant choice, while \textit{Pitch} plays a significant but secondary role. \textit{Pain Sound} (speaker identity) showed no significant influence, confirming that participants prioritized the intensity of the physical and auditory cues over the specific voice identity.

\vspace{3mm}

As shown in Figure~\ref{fig:ForceFreq}B, participant responses form overlapping but non-random clusters in PCA space. Notably, data points associated with choice `b' (agreement) appear more densely centered near the origin, suggesting that moderate values of features are generally associated with perceptual agreement. In contrast, choice `a' (strong agreement) and choice `d' (strong disagreement) tend to appear at opposite poles along PC1. This spatial distribution supports the hypothesis that the combination of sound intensity and force magnitude contributes more to consistent perceptual judgments than sound identity alone.

\vspace{3mm}

These findings indicate that perceived congruence in this study was heavily influenced by the intensity of the interaction (Amplitude and Force). Unlike the expectation that Pitch would be the primary driver of urgency, our data indicates that participants prioritized Force and Amplitude when determining the realism of the simulation, with Pitch playing a secondary, nuanced role. Speaker identity (Pain Sound) exerted negligible effect. These insights offer practical implications for robopatient design: to elicit strong user agreement, the system must prioritize synchronizing sound amplitude with applied force, while pitch variation may act as a subtle, rather than dominant, cue.


\subsection{Poststudy Questionaire}

The questionnaire was developed based on tailored questions specific to the study's objectives to cover the qualitative aspects of the experiment. The questionnaire was reviewed by subject matter experts to ensure content validity. Participants provided informed consent before the experiments.

\vspace{3mm}

We presented  7 Likert scale questions, followed up by an open-ended question to obtain any additional thoughts on the experiment. 

\vspace{3mm}

As the first question, we asked if any part of the experiment was particularly challenging or difficult, with an option to pick four options ranging from `very difficult' to `very easy'. We conducted a two-tailed t-test of two samples with unequal variance on the male and female preferences for this question, and the two groups seem to have no significant difference in their preference based on gender (p=0.533 $>$ 0.05). Similarly, the second question was to rate how comfortable they felt while performing the palpations. We did not observe any gender specific preferences in the answers for this either (p=0.146 $>$ 0.05). 

\vspace{3mm}

We further asked which attribute the participant paid attention to
first: facial expression or pain sound, or both, when they were performing palpations. We observed that 80\% of men paid attention to pain sound and 10\% to facial expressions first, and only 10\% for both, while only 60\% of women paid attention to both and 40\% to only pain sound. Overall 70\% of the participants paid attention to just the pain sound when palpating the robopatient.

\vspace{3mm}

The next question was to investigate if participants were able to distinguish between different levels of pain intensity based on sound, and 80\% recorded either ``distinguishable'' or ``very distinguishable''. Similar to previous responses, there was no significant difference between the two genders in perceiving the pain intensity in the form of sound (p=0.777 $>$ 0.05). As the next question, we asked if the participant feel any empathy towards the person you were palpating based on the pain sounds, and only 25\% of the participants responded "yes'' or "very much''. There was no significant difference in the preferences based on gender for this question either (p=0.423 $>$ 0.05). We then asked if they found it easier or harder to focus on the task as the experiment progressed to gain insights into the design of the experiment. 45\% of the participants responded that the experiment was getting easier over time while 55\% had an opposite impression. Based on the two-tailed t-test performed on the responses to this question, there was no gender-based significance in these preferences (p = 0.343 $>$ 0.05). Finally, we asked if the participant recollected any memories of their own experience about an encounter with a doctor in the past during the experiment, and 30\% responded ``yes'' to this question, and choices were equally distributed among the two genders.

\vspace{3mm}

Qualitative post-experiment feedback revealed that participants tended to focus more on vocal pain expressions due to their discrete nature.  The range of maximum pain intensity associated with vocal expressions may also contribute to this outcome, as participants might rely more on vocal cues to interpret pain levels. 

\vspace{3mm}

Participants suggested enhancing the robopatient simulator by adding diverse vocal pain expressions, such as grunting, and by developing more comprehensive palpation training programs. These programs should include scenarios for superficial and deep palpation, as well as training for identifying rebound tenderness.

\vspace{3mm}

Under additional comments from the participants, were suggestions to incorporate the speed of palpation to map appropriate pain sounds and include a more diverse set of pain sounds.  Furthermore, some commented that it was easier to distinguish female pain sounds compared to those of a male, and it would have been useful to know more about a certain injury and the type of pain to have more resemblance to the scenario. 

\subsection{Limitations, Implications and Future Work}
While promising, limitations such as a small sample size and restricted pain-to-sound mappings underscore the need for further research. Future work will expand participant diversity, incorporate medical professionals, and explore adaptive physical pain characteristics to improve training systems’ realism and efficacy.

\vspace{3mm}

Furthermore, the pain-to-sound mapping was limited to three levels, making it difficult for participants to differentiate feedback at maximum pain intensity. Additionally, the small sample size of naive participants may have reduced the statistical power, resulting in non-significant outcomes. This study is presented as a pilot to demonstrate the integration of vocal pain feedback into robopatients for abdominal palpation training. We plan to expand our investigations with a larger number of participants from diverse genders and ethnic backgrounds, and participants with different palpation experience levels, to address these limitations. In addition, we intend to expand the experiments to medical professionals. Furthermore, time scales of pain sound is also important to consider, for instance, higher pitch and louder pain cries are associated with a sense of urgency \cite{porter1986neonatal, craig1988judgment}. Hence, these properties will be incorporated into future pain expression generation.

\vspace{3mm}

The current approach relies primarily on explicit mappings of sound properties to force levels, which require a predefined framework for interpretation. To improve adaptability and intuitiveness, incorporating physical pain characteristics—such as surface compliance or deformation patterns—could allow the system to encode and process information directly through physical interaction. This would reduce reliance on predefined mappings and enable more seamless real-time adjustments based on the interaction itself.

\vspace{3mm}

Alongside pain-to-sound mapping, we will also improve the physical appearance of the robopatient to resemble a real patient lying on a hospital bed. Mainly, we will improve the design of the abdominal phantom by adding details such as the navel and abdominal muscles. In addition, better integration with MorphFace will improve the participant’s perception and empathy during palpation tasks and foster a more realistic interaction with the robopatient.

\section{Conclusions}\label{conclusion}

This study examined how acoustic features, specifically amplitude and pitch, interact with user-applied force thresholds to shape perceptual judgments of congruence in a palpation training context. Through a comprehensive analysis, we observed that participant agreement was primarily modulated by the intensity of the interaction, specifically the combination of auditory amplitude and tactile force, rather than the identity of the pain sound. Key findings from ANOVA and post-hoc tests revealed that amplitude and force threshold are the central drivers of perceived pain congruence in robot-mediated palpation, with louder sounds and higher force levels significantly linked to `strongly agree' responses. Pitch played a significant but secondary role, while pain sound identity (male or female voice) showed no statistically significant influence, indicating that speaker characteristics were not perceptually salient in this context.

\vspace{3mm}

Visualization in PCA space supported these statistical findings, revealing that participant choices form a gradient based on stimulus magnitude rather than distinct, isolated clusters. The data indicates that users rely on a magnitude-based heuristic to judge realism: high-intensity cues (high force paired with high amplitude) were most consistently linked to strong agreement. Conversely, mismatches in intensity (e.g., high force with low sound) led to disagreement.

\vspace{3mm}

The auditory pain expressions of robopatients can be optimized by altering the pitch and amplitude dynamically based on the palpation forces, with the specific audio having minimal significance. In our future work, we plan to conduct a large-scale study to further investigate and reinforce the findings of this work, and improve the robopatient to test in a clinical training environment.

\vspace{3mm}

As a whole, our results point to a perceptual framework where pitch and amplitude serve as core dimensions in shaping the realism and acceptability of robotic auditory feedback in simulated palpation tasks. This has direct implications for the design of robopatient systems, emphasizing the need for fine-grained acoustic modulation over simplistic force-to-sound mappings or reliance on speaker identity. For training scenarios that rely on simulating affective states such as pain, these insights can guide the development of more intuitive, convincing, and effective human-robot interaction systems.

\medskip
\textbf{Supplementary Materials}\par
The supplementary materials for this study are available on our \href{https://github.com/nsaitarun-git/Submission_Codes.git}{\color{blue}GitHub repository}. 
\medskip

\textbf{Ethics Statement}\par
The studies involving humans were approved by the Ethics Committee of the Department of Engineering, University of Cambridge, United Kingdom, under the light-touch/low-risk scheme (No.159, 14/04/2023). The studies were conducted by the local legislation and institutional requirements. The participants provided their written informed consent to participate in this study.
\medskip

\textbf{Conflict of Interest}\par
The authors declare no conflict of interest.

\medskip
\textbf{Acknowledgements} \par 
The authors are thankful to the participants who generously volunteered for this study. Each participant received a £10 Amazon voucher as a token of appreciation. This work was partially supported by the School of Engineering and Materials Science (SEMS) at Queen Mary University of London start-up funding and the European Union’s Horizon
2020 research and innovation programme under the Marie Skłodowska-Curie grant agreement No 101034337.

\medskip
\textbf{Data Retention} \par

We collected force data from the trials, the age, and the gender of each participant. The responses to the experiment and the post-study questionnaire were kept anonymous to ensure participant confidentiality.

\medskip

\bibliographystyle{MSP}
\bibliography{References}

\end{document}